\pdfoutput=1

\documentclass[11pt]{article}

\usepackage[final]{acl}

\usepackage{times}
\usepackage{latexsym}

\usepackage[T1]{fontenc}

\usepackage[utf8]{inputenc}

\usepackage{microtype}

\usepackage{inconsolata}

\usepackage{graphicx}

\usepackage{pifont}

\usepackage{amsthm,amsmath,amssymb}
\usepackage{multirow} 
\usepackage{tcolorbox}
\usepackage{hyperref}
\usepackage{enumitem}
\usepackage{multicol}
\usepackage{booktabs} 
\usepackage{fontawesome}
\usepackage{array}
\usepackage{colortbl} 
\definecolor{lightblue}{RGB}{192,224,255} 
\title{\textit{iTool}: Reinforced Fine-Tuning with Dynamic Deficiency Calibration for Advanced Tool Use}

\author{
 \textbf{Yirong Zeng \textsuperscript{1}},
 \textbf{Xiao Ding\thanks{Corresponding author. Email: xding@ir.hit.edu.cn} \textsuperscript{1}},
 \textbf{Yuxian Wang\textsuperscript{2}},
 \textbf{Weiwen Liu\textsuperscript{3}},
 \textbf{Wu Ning\textsuperscript{2}}, \\
 \textbf{Yutai Hou\textsuperscript{2}},
 \textbf{Xu Huang\textsuperscript{4}}, 
 \textbf{Duyu Tang\textsuperscript{2}},
 \textbf{Dandan Tu\textsuperscript{2}},
 \textbf{Bing Qin\textsuperscript{1}},
 \textbf{Ting Liu\textsuperscript{1}},
\\
 \textsuperscript{1}Harbin Institute of Technology SCIR Lab,
 \textsuperscript{2}Huawei Technologies Co., Ltd, \\
 \textsuperscript{3}Shanghai Jiao Tong University,
 \textsuperscript{4}University of Science and Technology of China \\ 
\\
}

\begin{document}
\maketitle
\begin{abstract}
Augmenting large language models (LLMs) with external tools is a promising approach to enhance their capabilities, especially for complex tasks.
Synthesizing tool-use data through real-world simulations is an effective way to achieve this.
However, our investigation reveals that training gains significantly decay as synthetic data increases.
The model struggles to benefit from additional synthetic data, which fails to endow it with advanced tool-use capabilities in complex scenarios
Moreover, we discovered that the above limitation usually manifests as a fragment deficiency (i.e., parameter errors) in response.
To this end, we propose an iterative reinforced fine-tuning strategy designed to alleviate this limitation.
This strategy involves: (1) enhancing the diversity of response for synthetic data through path exploration of Monte Carlo Tree Search.
(2) iteratively pinpointing the model's deficiency by constructing fine-grained preference pairs, and then improving it by preference optimization algorithms for targeted improvement.
The experiments show that our method achieves 13.11\% better performance than the same-size base model. It achieves an improvement of 6.5\% in complex scenarios compared to the baseline, and it also outperforms larger open-source and closed-source models\footnote{Code: \url{https://github.com/zeng-yirong/iTool} }.

\end{abstract}

\section{Introduction}
Integrating LLMs with external tools significantly enhances their capability to tackle complex tasks in real-world scenarios \citep{li2025review, qu2024tool}. 
For instance, the tool-use capability allows LLMs to access up-to-date information, perform precise calculations, and reduce the likelihood of hallucinations \citep{singh2025agentic}. 
This unlocks a wide range of potential applications in various domains, such as complex reasoning tasks \citep{li2025adaptive,manduzio2024improving}, and the scheduling of applications on devices \citep{gunter2024apple,luo2025self}.
In essence, tool use involves the following process: Given one or more tools, a user presents a question, and the LLM selects the appropriate tools from the candidate tools and performs the tool call to fulfill the user's demands. 
In this paper, \includegraphics[height=11pt]{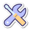} tools are used interchangeably with APIs, functions, and plugins.

\begin{figure}[t]
    \centering
  \includegraphics[width=0.99\linewidth]{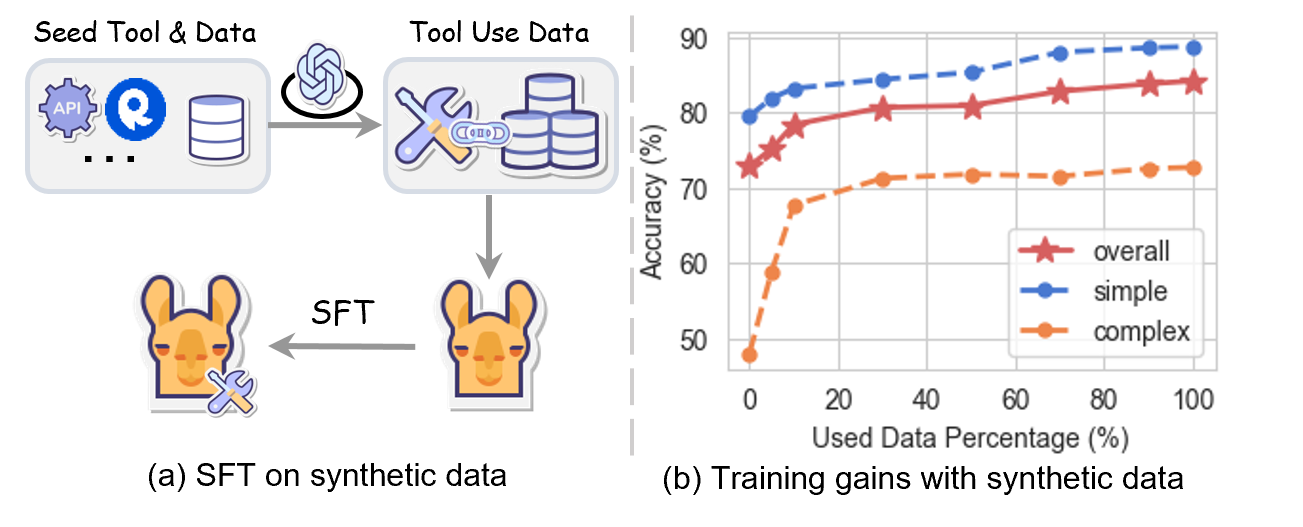}
  \caption{The training paradigm of the tool-use model under synthetic data (a). 
  However, as shown in (b), the growth rate of the model’s performance gain declines significantly as the training data increases, especially in complex tool-use scenarios. }
  \label{fig:previous_work}
\end{figure}

Recent advancements have found that LLMs can handle simple tool use scenarios through prompt engineering \citep{ye2024tl}, but they encounter difficulties with more complex real-world applications (e.g., long contexts or extensive toolsets) \citep{bfclv3}.
To address this, some studies simulate real-world scenarios, such as ticketing systems, to mimic more realistic use cases \citep{lin2024hammer} to collect synthetic data. 
Synthetic data are used in supervised fine-tuning (SFT) to improve tool use in complex scenarios, as shown in Figure \ref{fig:previous_work} (a).
Despite these solution strides in the development of tool-use models, our investigation reveals a critical weakness: there is a training gains decay as the synthetic tool-use data scales.

We conducted tests to explore how the performance of the model changes when synthetic data of different proportions is used, as shown in Figure \ref{fig:previous_work} (b), 
We find that the model struggles to benefit from more synthetic data with SFT in complex scenarios. 
More analysis in Section \ref{sec:pre_study} indicates that this limitation reflects the failure of the model to extract the parameter name or infer the correct parameter value from the user query. 
This issue typically affects only a small fragment of the response, differing from the ground truth response.


Therefore, we attempt to alleviate the decay of training gains when using synthetic tool-use data, to enhance the ability of tool use in complex scenarios.
It is not easy because it requires equipping the model with advanced contextual understanding and reasoning capabilities.
Fortunately, the success of OpenAI o1\footnote{https://openai.com/index/learning-to-reason-with-llms/} demonstrates complex reasoning through step-by-step slow thinking (e.g., Monte Carlo Tree Search (MCTS) \citep{coulom2006efficient} ) and Reinforced Fine-Tuning (ReFT) \citep{luong2024reft} (tailors reinforcement learning and aligns with user intentions to specific tasks). 

To this end, we propose a novel learning method involving (1) an MCTS-based path exploration to enhance response diversity and (2) ReFT to progressively correct the wrong fragment text of model's response.
Specifically, we propose an \underline{i}terative reinforced fine-tuning strategy for \underline{Tool} use, named \textit{iTool}. 
It first iteratively identifies complex data based on feedback from a policy model. 
It then performs MCTS to help explore data diversity in response, and further pinpoint wrong fragment by collecting fine-grained preference pairs from search path.
Finally, a reinforcement learning policy (i.e., direct preference optimization \citep{rafailov2024direct}) is applied to align the model's response with the ground-truth response and misalign it with wrong fragment.
Moreover, before iterative ReFT, we propose an easy-to-hard warm-up SFT strategy for better learning from complex scenarios.
Following these advancements, \textit{iTool} demonstrates \textasciitilde13\% better performance than the base model.
It also achieves substantial improvements in tool-use ability under complex scenarios.
Despite having only 8B parameters, it outperforms larger open-source models and competes with top-tier closed-source models.


\section{Problem Statement and Analysis}
\label{sec:motivation}
\subsection{Task Overview}
\begin{figure}[t]
    \centering
  \includegraphics[width=1.0\linewidth]{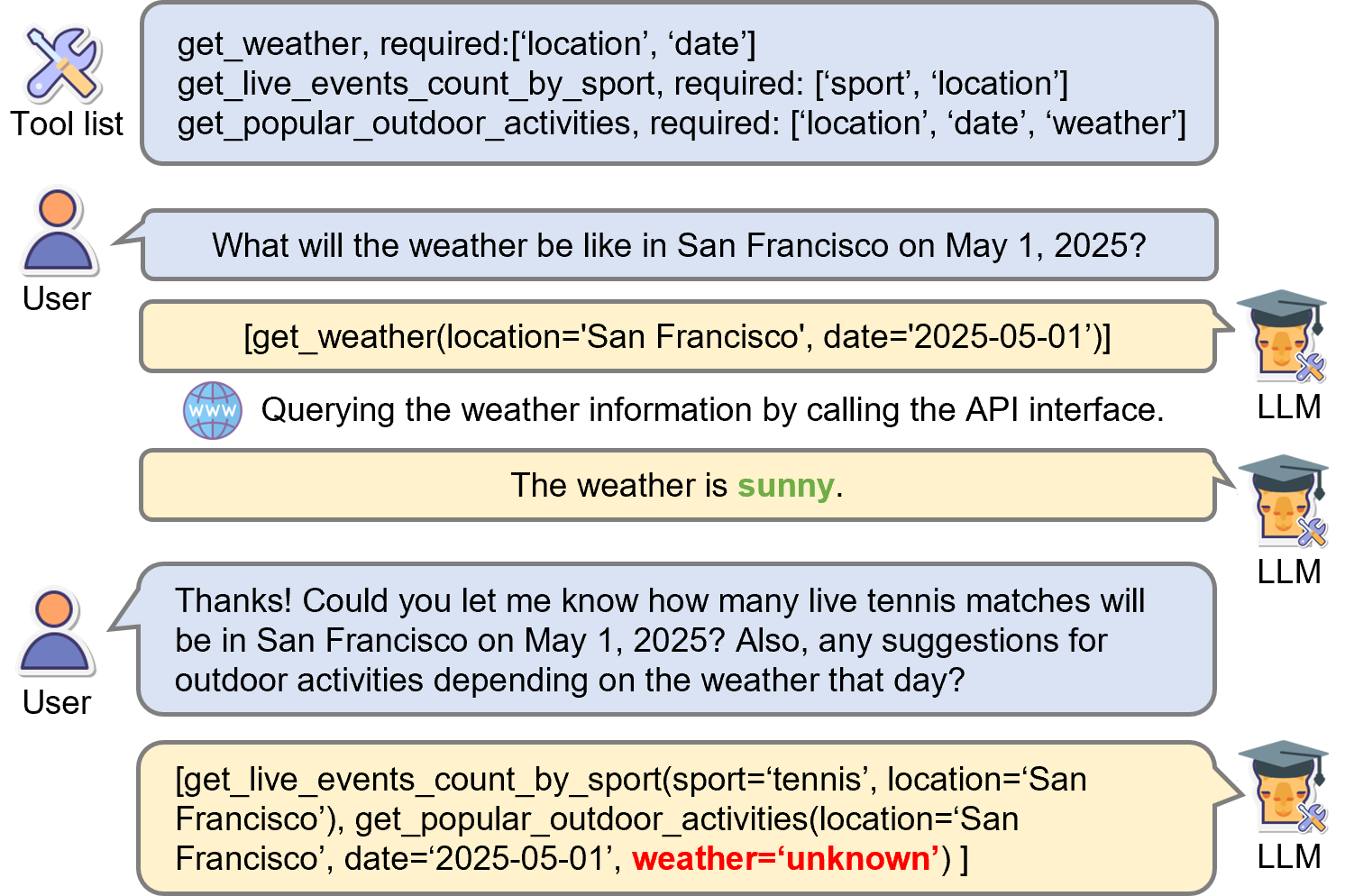}
  \caption{An illustration of tool-use. Given a user query with candidate tools, LLMs select the tool(s) from candidates, then execute the API call operation, and finally reply with a response.
  In the bad response, the parameter errors (i.g, red font \texttt{weather='unknown'}) account for a small fragment of the response content.
  }
  \label{fig:example_of_tool_use}
\end{figure}

 In tool use, the LLM receives a user query \( q \) along with a set of candidate tools, represented as \( \mathcal{T} = \{ {t}_0, {t}_1, \dots, {t}_{|\mathcal{T}|} \} \). 
 The purpose of LLM is to fulfill the user’s intent by executing a specific sequence of tools. 
 The decision process can be described as \( y \sim \pi(y \mid s_0, q, \mathcal{T} ) \), where \( \pi(\cdot) \) represents the policy model, \( s_0 \) denotes the initial task state, and \( y \) represents the actions taken by the model, such as selecting or executing a specific tool call from \( \mathcal{T} \). 
A case is illustrated in Figure \ref{fig:example_of_tool_use}.

\subsection{Preliminary Study}
\label{sec:pre_study}
This section presents the challenges when fine-tuning models with tool-use synthetic data, and clarifies the motivation for the proposed methods.

We fine-tune the model using synthetic tool-use data of varying proportions. 
Specifically, training data: \textbf{ToolACE} \citep{liu2024toolace} is a general tool-use dataset with up to 100K samples, and created through a novel self-evolution synthesis.
Evaluation benchmark: Berkeley Function-Calling Leaderboard (\textbf{BFCL}) \citep{bfclv3} provides a comprehensive dataset comprising 4k+ instances (updating), consisting of \textit{Non-live} (with expert-curated simple tools), \textit{Live} (with user-contributed complex tools),  \textit{Multi-turn} (with multi-turn \& multi-step tool use) and Hallucination (i.e., relevance and irrelevance detection) samples.
Here, \textit{Non-live} denotes {simple} tool use scenarios (e.g., single tool), while \textit{Live} represents more {complex} tool use scenarios (e.g., multiple parallel tools). 
For convenient understanding, in this section, we use \textbf{simple} and \textbf{complex} as aliases for the \textit{Non-live} and \textit{Live} metrics, respectively.

The results are depicted in Figure \ref{fig:previous_work} (b). 
We observe that the model's performance gain declines significantly as the training data increases. 
Specifically, with the SFT paradigm shown in Figure \ref{fig:previous_work} (a), 
The model significantly enhances tool-use ability with small-scale supervised data by mimicking patterns from the training examples.
However, the performance improvement significantly declines after 30\% of the data is used.
The model struggles to benefit from using more synthetic data, we argue that insufficient data diversity is one of the key factors. 

\begin{figure}[t]
    \centering
  \includegraphics[width=0.95\linewidth]{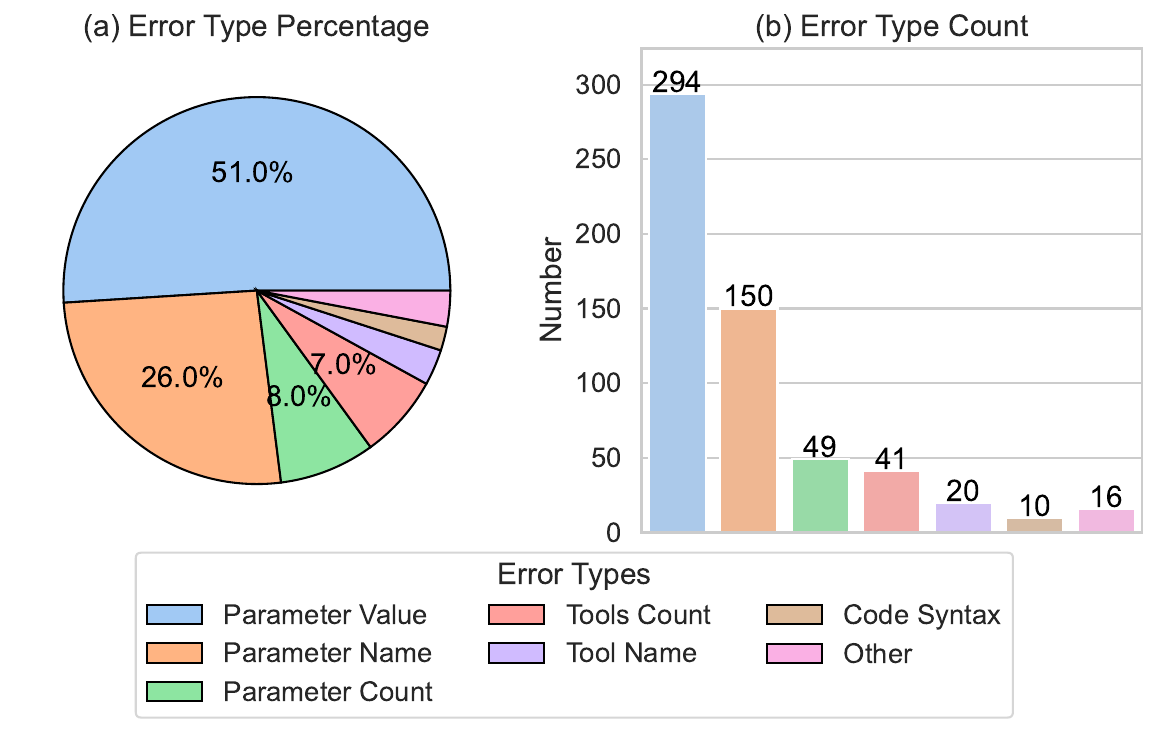}
  \caption{ Error type distribution in bad cases. 
  In bad cases, error types are highly concentrated in \textit{Parameter Value \& Name}.
  }
  \label{fig:bad_case_idd}
\end{figure}

To explore the manifestations of the above-mentioned issue, we perform a bad case analysis.
We counts all error types in \textit{Live} and \textit{Non-live} of BFCL, and categorized the error types as shown in Figure \ref{fig:bad_case_idd}.
Here, \textit{Parameter Value} error denotes the value of the parameter that does not match the ground truth.
\textit{Parameter Name} error denotes unable to identify the parameter value from the user query.
For more details, see Appendix \ref{sec:error_types}.
From Figure \ref{fig:bad_case_idd}, we observed that errors are highly concentrated in \textit{Parameter Value \& Name} errors. 
In bad cases, parameter error constitutes a small fragment in response, while the majority remains consistent with the ground-truth.
An illustration is shown in Figure \ref{fig:example_of_tool_use}.
Therefore, trying to fix the fragment error can help alleviate the limitation of gain decay in training models.

In summary, we find that training with synthetic tool-use data causes gain decay, and the model struggles to benefit from additional such data.
This limitation is reflected in the model’s deficiency (i.e., parameter errors) in responses.
Motivated by this line, we utilize the MCTS path to explore diversity in responses for alleviating such gains decay.
We further propose an iterative ReFT strategy to progressively pinpoint and optimize the model's deficiencies.

\section{Method}
\begin{figure*}[t]
    \centering
  \includegraphics[width=0.85\linewidth]{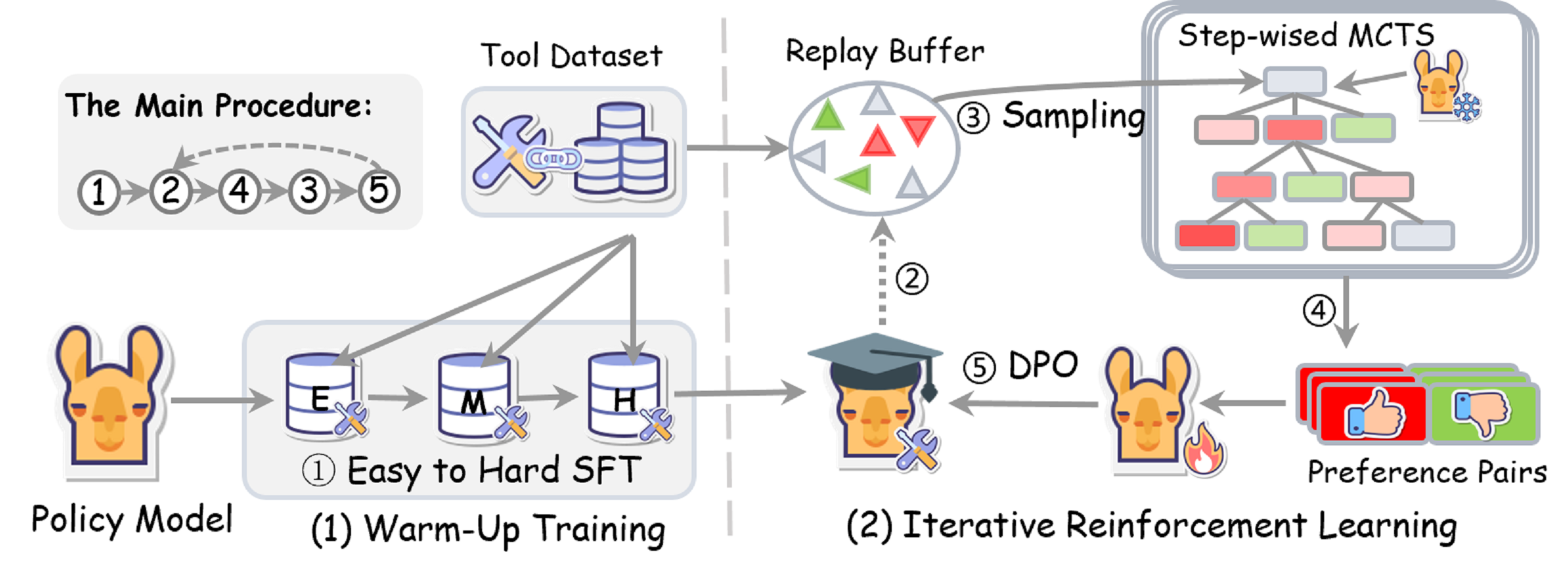}
  \caption{The overall architecture of \textit{iTool} consists of warm-up training and iterative reinforcement learning. 
  Specifically, after warm-up training \ding{172}, the policy model refreshes the replay buffer \ding{173} and then actively samples complex data \ding{174}. 
  Then, step-wise MCTS \ding{175} is performed to obtain fine-grained preference pairs for pointing out the wrong fragment in response. 
  Finally, the models are updated via direct preference optimization \ding{176} to improve response. 
  The fire \includegraphics[height=9pt]{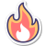} and frozen \includegraphics[height=9pt]{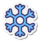} denote parameters are updated and fixed, respectively.
  }
  \label{fig:main}
\end{figure*}

In this section, we provide a detailed introduction to our method.
Figure \ref{fig:main} shows the overall architecture.
It consists of warm-up training and iterative reinforcement learning.

\subsection{Warm-up training}
In real-world applications, the tool-use model should select multiple tools from a complex candidate toolset and schedule them correctly (a.k.a., hard mode), instead of directly using a single candidate tool to respond (a.k.a., easy mode). 
Similar to human learning procedures, tool learning models can benefit from an easy-to-hard curriculum during model training \citep{xu2020curriculum}.
Therefore, we propose an easy-to-hard SFT for warm-up training. 

In the warm-up stage, we first divide the dataset evenly into three subsets (i.e., easy, medium, hard) based on difficulty levels.
We follow the criteria: (a) the candidate toolset number; (b) the string length of the toolset; and (c) the number of tool calls needed in response to split the dataset.
The specific definitions for each subset are as follows:
(1) hard: \texttt{a >= 4 or b > 2000 or c >= 4}.
(2) medium: \texttt{1 < a < 4 or b < 2000 or c < 4}.
(3) simple: \texttt{a <= 1 and b < 1000 and c <= 1}. 
\begin{equation}
\mathcal{D} = \mathcal{D}_{easy} \bigcup \mathcal{D}_{medium} \bigcup \mathcal{D}_{hard} .
\end{equation}


Subsequently, we fine-tune the LLM \( \mathcal{M} \) sequentially on each subset \( \mathcal{D}_i \) using the supervised loss:
\begin{equation}
\mathcal{L}_i = -\mathbb{E}_{(q,y) \sim \mathcal{D}_i} \left[ \log P_{\mathcal{M}}(y \mid q, \mathcal{T}) \right],
\end{equation}
with \( \mathcal{D}_1 \) (easy), \( \mathcal{D}_2 \) (medium) and \( \mathcal{D}_3 \) (hard).

The total warm-up loss is:
\begin{equation}
\mathcal{L}_{\text{warm-up}} = \sum_{i=1}^{N=3} \mathcal{L}_i.
\end{equation}

\subsection{MCTS-Based Iterative Reinforcement Learning}
\label{sec:method}
In order to alleviate training gains decreases using synthetic tool-use data for LLM, in this module, we propose an Iterative Reinforcement Learning scheme to continuously remedy this deficiency.
As shown in Figure \ref{fig:main}, it iteratively refreshes replay buffer to sample complex data and generates preference data for preference optimization.

\textbf{Sampling complex data}. 
Given a warm-up model from the previous stage, it is used to refresh the replay buffer by feeding back the complexity of samples.
The replay buffer is initialized with a random 50\% sample from the tool-use dataset.
Each example in the buffer is represented as:
\( x_{buff} = \langle {q}, \mathcal{T}, {c} \rangle \), where \( c \) is denote the complexity of sample.
In practice, model generation perplexity \( h \) is used to measure the complexity of the samples, i.e., \( c = h\).
The generation perplexity of the target response can be factorized as follows:
\begin{equation}
h = \sqrt[n]{\frac{1}{P_{\mathcal{M}}(y \mid q, \mathcal{T})}},
\end{equation}
where the \( P_{\mathcal{M}}(y \mid q, \mathcal{T}) \) is the generation probability.
Since perplexity \( h \) represents the degree of generation uncertainty \citep{gao2024confucius}, we sample top 10\% highest \( h \) data for subsequent step in each iteration.

\textbf{MCTS for Step-Level Preference. }
The success of OpenAI o1 provides a compelling illustration of the effectiveness of step-by-step thinking.
As a key algorithm, MCTS path exploration can fully traverse the search space and provide greater data diversity \citep{grill2020monte}.
Inspired by these, we propose to integrate MCTS into training for collecting step-level preference data.

The step-wise MCTS is achieved by breaking down the expansion step into discrete steps, transforming instance-level rewards into granular step-level signals.
Specifically, it begins from a root node \( s_0 \) (i.e., user query), and unfolds in three iterative stages: selection, expansion, and backup:

\noindent
\textbf{ (1) Select.} It is guided by two key variables: \( Q(s_t, a) \) is the value of taking action \( a \) in state \( s_t \), and \( N(s_t) \) is the visitation frequency of state \( s_t \).
We employ the Predictor+ Upper Confidence bounds applied to Trees (PUCT) \citep{rosin2011multi} to navigate the trade-off between exploring and exploiting ones. 
At node \( s_t \), the subsequent node follows the formula:
\begin{equation}
\small
s_{t+1} = \arg\max_{a}\left[ Q(s_t, a) + c \cdot p(a \mid s_t) \frac{\sqrt{N(s_t)}}{1 + N({n}(s_t, a))} \right]
\end{equation}
where \( p(a\mid s_t) = \pi_\theta(a \mid q, \mathcal{T}, s_t) \) denotes the policy \( \pi_\theta(\cdot) \)'s probability distribution for generating a action step \( a \), and \( c \) is the trade-off hyperparameter, and \({n}(s_t, a)\) explicitly represents the next state generated by taking action \( a \) in state\(s_t\).
We enforce the policy model to generate fine-grained fragments (e.g., an argument assignment operation, like \texttt{weather=’unknown’} in Figure \ref{fig:example_of_tool_use}) by managing the termination characters (e.g., \texttt{`,. )'}).

\noindent
\textbf{ (2) Expand}. It occurs at a leaf node during the selection process to integrate new nodes and assess rewards. The reward \( r(s_t, a) \) for executing step \( a \) in state \( s_t \) is quantified by the reward difference between states \( \mathcal{R}(s_t) \) and \( \mathcal{R}(s_{t+1}) \), showing the benefit of action \( a \) in state \( s_t \).
As defined in Eq.\ref{reward}, reward computation merges outcome correctness \( \mathcal{O} \) with self-evaluation \( \mathcal{C} \). 
Following \citet{xie2024monte}, we define self-evaluation with Eval Prompt \ref{eval_prompt}  as Eq.\ref{eval}.
\begin{equation}
\mathcal{R}(s_t) = \mathcal{O}(s_t) + \mathcal{C}(s_t) \label{reward},
\end{equation}
\begin{equation}
\mathcal{C}(s_t) = \pi_\theta( cs \mid {prompt}_{{eval}}, q, a, \mathcal{T}, s_t) \label{eval},
\end{equation}
 where \( cs \) denotes the \underline{c}onfidence \underline{s}core in token-level probability for correctness. 
Future rewards are anticipated by simulating upcoming scenarios through roll-outs, following the selection and expansion process until reaching a terminal state (i.e., complete response or exceeds the maximum length).

\noindent
\textbf{(3) Backup.} 
Once a terminal state is reached, we carry out a bottom-up update from the terminal node back to the root. 
We update the visit count \( N \), the state value \( V \), and the action value \( Q \):

\begin{equation}
\small
V(s_t) \leftarrow \sum_a N(s_{t+1}) Q(s_t, a) / \sum_a N(s_{t+1}),
\end{equation}
\begin{equation}
Q(s_t, a) \leftarrow r(s_t, a) + \gamma V(s_{t+1}),
\end{equation}
where \( \gamma \) is the discount for future state values.

We use the action value \( \mathcal{Q} \) to indicate the preference for candidate steps, with higher values showing more preferred next steps.
For each node in the search tree, we choose the steps with the highest and lowest \( \mathcal{Q} \) as the preferred and dispreferred responses, respectively, and consider the prefix path as the question. 
See Appendix \ref{example_pair} for an example.
Therefore, our method leverages MCTS to generate numerous negative trajectories with fine-grained deficiencies, thereby enhancing data diversity.

\textbf{Iterative preference optimization.}
Given the step-level preferences collected via MCTS, we tune the policy model via SimPO \citep{meng2024simpo}, a variant of DPO \citep{rafailov2024direct},
because it reduces computational overhead by eliminating the need for a reference model.
After optimization, we obtain the updated policy \(\pi_{\theta(i)}\) and repeat sampling the complex data process to iteratively update the policy model. 

As a variant of DPO, it eliminates the need for a reference model and introduces a simple reference-free reward aligned with generation, i.e., length-normalized reward:
\begin{equation}
r_{\text{SimPO}}(x, y) = \frac{\beta}{|y|} \sum_{i=1}^{|y|} \log \pi_\theta(y_i \mid x, y_{<i}),
\end{equation}
where $\beta$ is a constant that controls the scaling of the reward difference.
Using the shorthand \( h_{\pi_{\theta}}^{y_{w}} = \frac{\beta}{|y_w|}\log \pi_\theta(y_w | x), h_{\pi_{\theta}}^{y_{l}} = \frac{\beta}{|y_l|} \log \pi_\theta(y_l | x)\),
at the \(i\)-th iteration, given a batch of preference data \(\mathcal{D}_{i}\) sampled with the latest policy \(\pi_{\theta(i-1)}\), we denote the policy objective \(\ell_{i}(\theta)\) as follows:
\begin{equation}
\small
\ell_{i}(\pi_\theta) = -\mathbb{E}_{(x, y_w, y_l) \sim \mathcal{D}_{i}} \left[ \log \sigma \left(  h_{\pi_{\theta}}^{y_{w}} - h_{\pi_{\theta}}^{y_{l}}- \gamma \right) \right],
\end{equation}
where \( \gamma>0 \) represents the target reward margin, ensuring that the preferred response's reward exceeds that of the dispreferred one; \(y_{w}\) and \(y_{l}\) represent the step-level preferred and dispreferred responses, respectively.


\section{Experiments}
\subsection{Experimental Setup}
\label{exp_setup}
We take the widely used open-source LLM, LLaMA3.1-8B-Instruct
as our base model.
We use synthetic data from ToolACE for experiments, randomly select 90\% for warm-up training, and 50\% for reinforcement learning to balance performance and cost.
For warm-up training, we adopt the parameter-efficient training strategy LoRA \citep{hu2022lora}.
For reinforcement learning, we employ SimPO, a variant of DPO, for preference optimization, utilizing the QLora parameter-efficient training strategy \citep{dettmers2024qlora}.
For more implementation details and preferences optimization analysis, see Appendix \ref{sec:appendix}.

\noindent
\textbf{Evaluation Dataset}. 
In addition to \textbf{BFCL}, we use \textbf{API-Bank} \citep{li2023api}, which consists of 314 tool-use dialogues and 753 API calls. 
This dataset evaluates models' abilities to correctly invoke a known API (L-1) based on a query and to retrieve and call APIs from a tool list (L-2).

\noindent
\textbf{Baselines}
We compare the overall performance with the state-of-the-art closed-source models (e.g., GPT-series, Gemini  and open-source models (e.g., Llama-3.1-8B-Instruct, Qwen2.5-7B \citep{qwen2.5}), as well as fine-tuned open-source models with tool-use dataset, including ToolACE-8B (fine-tuning Llama-3.1-8B-Instruct on ToolACE) model, xLAM-series \citep{zhang2024xlam} and Hammer-series \citep{lin2024hammer}.

\begin{table*}[th]
    \small
    \centering
    \begin{tabular}{cc|l|ccc|c}
        \toprule
        Rank & Overall Acc & Model & Non-live & Live  & Multi turn & Rel / Irrel \\
        \hline  \midrule
\rowcolor{lightblue} 1 & 63.26 & \(\clubsuit\) \textbf{iTool-8B (FC)} & 88.82 & \textbf{78.29} & 23.84 & 84.90/80.72 \\
        2 & \textbf{62.19} & \(\spadesuit\) GPT-4o-2024-08-06 (FC) & 86.15 & 75.43 & \textbf{25.00} & 63.41/82.93 \\
        3 & 61.89 & \(\spadesuit\) GPT-4-turbo-2024-04-09 (FC) & 88.80 & 76.23 & 24.88 & 73.17/79.76 \\
        4 & 60.47 & \(\spadesuit\) GPT-4o-mini-2024-07-18 (FC) & 83.72 & 70.19 & 27.50 & 80.49/71.77 \\
        5 & 60.44 & \(\clubsuit\) ToolACE-8B (FC) & {88.94} & 74.99 & 17.38 & 80.49/85.71 \\
        6 & 58.15 & \(\spadesuit\) GPT-4o-mini-2024-07-18 (Prompt) & 88.69 & 74.63 & 11.13 & 75.61/81.00 \\
        7 & 57.99 & \(\clubsuit\) xLAM-8x22b-r (FC) & 87.51 & 71.97 & 14.50 & 85.37/67.29 \\
        8 & 57.92 & \(\spadesuit\) Gemini-1.5-Flash-002 (Prompt) & 87.60 & 76.28 & 9.88 & 85.37/78.54 \\
        9 & 57.69 & \(\clubsuit\) Hammer2.0-7b (FC) & 88.54 & 69.79 & 14.75 & 95.12/68.46 \\
        10 & 57.45 & \(\spadesuit\) o1-mini-2024-09-12 (Prompt) & 83.84 & 75.39 & 13.12 & 48.78/88.04 \\
        11 & 56.80 & \(\heartsuit\) mistral-large-2407 (FC) & 81.41 & 68.37 & 20.62 & 75.61/49.44 \\
        12 & 56.51 & \(\spadesuit\) Gemini-1.5-Pro-002 (Prompt) & \textbf{89.63} & 74.41 & 5.50 & 65.85/77.30 \\
        13 & 55.86 & \(\spadesuit\) Gemini-1.5-Flash-001 (Prompt) & 85.74 & 69.21 & 12.62 & 82.93/67.84 \\
        14 & 55.78 & \(\spadesuit\) GPT-4-turbo-2024-04-09 (Prompt) & 88.80 & 69.04 & 9.50 & 82.93/58.95 \\
        15 & 55.10 & \(\spadesuit\) Gemini-1.5-Pro-001 (Prompt) & 86.17 & 73.12 & 6.00 & 56.10/85.00 \\
        16 & 54.41 & \(\clubsuit\) xLAM-7b-r (FC) & 80.86 & 67.88 & 14.50 & 97.56/64.05 \\
        17 & 54.27 & \(\heartsuit\) Qwen2.5-7B-Instruct (Prompt) & 85.58 & 65.97 & 11.25 & 92.68/64.95 \\
        18 & 53.67 & \(\heartsuit\) Llama-3.1-70B-Instruct (Prompt) & 87.50 & 61.13 & 12.38 & 92.68/58.38 \\
        19 & 53.66 & \(\heartsuit\) Gemma-2-27b-it (Prompt) & 87.39 & 69.48 & 4.12 & 87.80/68.76 \\
        20 & 53.00 & \(\spadesuit\) GPT-3.5-Turbo-0125 (FC) & 78.52 & 61.22 & 19.25 & 97.56/35.16 \\ \hline
        21 & 52.50 & \(\heartsuit\) Gemma-2-9b-it (Prompt)	&84.52 &	69.21 &	3.75 &	87.80/72.45 \\
        22 & 51.59 & \(\clubsuit\) Hammer2.0-1.5b (FC) &	84.44 &	63.22 &	7.13 &	92.68/60.64 \\
        23 & 51.50 & \(\heartsuit\) Meta-Llama-3-70B-Instruct (Prompt) &	85.10 &	66.15 &	3.25 &	92.68/52.78 \\
        27 & 50.15 & \(\heartsuit\) Llama-3.1-8B-Instruct (Prompt)	& 81.15 &	57.93 &	11.38 &	78.05/41.62 \\
        28 & 49.02 & \(\clubsuit\) xLAM-8x7b-r (FC) &	73.93 &	69.12 &	4.00 &	87.80/68.12  \\
        29 & 48.82 & \(\heartsuit\) Qwen2.5-1.5B-Instruct (Prompt)	& 53.99 &	61.71 &	6.62 &	75.61/67.17 \\
        42 & 42.98 & \(\heartsuit\) Llama-3.2-3B-Instruct (Prompt)	& 11.11 &	50.91 &	4.00 &	63.41/68.81 \\
        \bottomrule
    \end{tabular}
    \caption{The leaderboard of different models in four tool-use scenarios of BFCL (v3) benchmark . 
    The top 20 models and baselines are listed for comparison. 
    FC denotes the model is tailored for functional calling. 
    Rel and Irrel denote relevance and irrelevance detection, respectively, indicating whether to call a tool or not.
     \(\spadesuit\) denotes closed-source model, \(\heartsuit\) denotes open-source base model, \(\clubsuit\) denotes open-source fine-tuned model.
    }
    \label{overall_res}
\end{table*}

\subsection{Overall Performance}
The overall performance of \textit{iTool-8B} and baseline models are shown in Table \ref{overall_res} and Table \ref{api_bank}.
Our model consistently achieves superior performance at comparable scales ($ \sim$ 8B).
Specifically, it shows consistent advantageous performance on API-Bank and BFCL compared with open-source models, and also outperforms most closed-source and larger open-source models in BFCL (e.g., GPT-4-series models).
For example, it outperforms xLAM-8x22b-r by 5.27 in the overall accuracy metrics.
Moreover, it demonstrates its superiority in challenging scenarios (e.g., \textit{Live}), which indicates our method learn advanced tool-use capabilities effectively from synthetic data.
This is primarily due to our iterative ReFT strategy, which continuously pinpoints and optimizes the model's deficiencies.

\begin{table}[th]
    \centering
    \small
    \begin{tabular}{l|cc}
         \toprule
         \multirow{2}{*}{Model}  & API-Bank & API-Bank \\ 
                                  &  L1  & L2   \\ 
        \hline \midrule
        {\(\spadesuit\)}  GPT-3.5-turbo-0125 & 70.43 & \textbf{52.59} \\  
        {\(\spadesuit\)} GPT-4-0613 & 75.94 & 48.89 \\  
        {\(\spadesuit\)} GPT-4-turbo-2024-04-09 & 72.43 & 39.26 \\  
        {\(\spadesuit\)} GPT-4o-mini-2024-07-18 & 74.69 & 45.93 \\  
        {\(\spadesuit\)} \textbf{GPT-4o-2024-05-13} & \textbf{76.19} & 42.96 \\ 
         \midrule
        {\(\heartsuit \)} Alpaca-7B & 24.06 & 5.19 \\  
         \(\heartsuit\) ChatGLM-6B & 23.62 & 13.33 \\  
         \(\clubsuit\) Lynx-7B & 49.87 & 30.37 \\  
         \(\clubsuit\) xLAM-7b-fc-r & 32.83 & 21.48 \\  
         \(\heartsuit\) LLaMA-3.1-8B-Instruct & 71.18 & 37.04 \\  
         \(\heartsuit\) Qwen2.5-7B-Instruct &  72.83 & 41.98 \\
         \(\clubsuit\) {ToolACE-8B} & {75.94} & {47.41} \\ 
         \(\clubsuit\) \textbf{iTool-8B} & \textbf{78.89} & \textbf{52.87} \\ 
         \bottomrule
    \end{tabular}
    \caption{Accuracy performance comparison on API-Bank evaluation system. \textbf{Bold} values represent the highest performance.}
    \label{api_bank}
\end{table}

\begin{table}[th]
    \centering
    \small
    \begin{tabular}{l|ccc}
        \toprule
        Models & Non-live & Live & Multi-turn \\ \midrule
        Base Model & 81.15 & 57.93 & 11.38 \\
        \hspace{0.2cm}\textit{+ base SFT} & 88.94 \textcolor{green!40!black}{\textsubscript{↑7.8}} & 74.99 \textcolor{green!40!black}{\textsubscript{↑17}} & 17.38 \textcolor{green!40!black}{\textsubscript{↑6.0}} \\
        \hspace{0.4cm} \textit{+ IRT} &	88.86 \textcolor{red!70!black}{\textsubscript{↓0.1}} &	76.51 \textcolor{green!40!black}{\textsubscript{↑1.5}}	& 20.65 \textcolor{green!40!black}{\textsubscript{↑3.3}} \\
        \hspace{0.2cm}\textit{+ warm-up SFT} & 88.35 \textcolor{green!40!black}{\textsubscript{↓7.2}} & 75.84 \textcolor{green!40!black}{\textsubscript{↑17.9}} & 19.65 \textcolor{green!40!black}{\textsubscript{↑8.3}} \\
        \hspace{0.4cm}\textit{+ IRL (iTool)} & 88.82 \textcolor{green!40!black}{\textsubscript{↑0.5}} & 78.29 \textcolor{green!40!black}{\textsubscript{↑3.2}} & 23.84 \textcolor{green!40!black}{\textsubscript{↑4.2}} \\ \midrule
        Total & \textcolor{green!40!black}{{↑9.5}} & \textcolor{green!40!black}{{↑21.2}} & \textcolor{green!40!black}{{↑12.5}} \\ \bottomrule
    \end{tabular}
    \caption{The module ablation performance (↑ = increase, ↓ = decrease). }
    \label{tab:ablation_1}
\end{table}

\subsection{Ablation  Analysis}
\subsubsection{Module Ablation}
To evaluate the effectiveness of the two components in our method, we conduct an ablation study in:
(1) the warm-up training phase (\textit{w/o warm-up}).
(2)  the Iterative Reinforcement Learning (IRL) module (\textit{w/o IRL}).
We adopt LLaMA-3.1-8B-Instruct as the \textit{Base} model for benchmarking, ensuring a consistent baseline across all experimental conditions.
From Table \ref{tab:ablation_1}, we find that all components are essential within our method.
\textit{base SFT} denotes SFT with the entire gold labeled dataset.
iTool achieves a comparable level to \textit{SFT} on the Non-live metric, but each module brings substantial improvements on the complex-scenario metrics (Live and Multi). 
Specifically, the warm-up training and IRL modules individually contribute improvements of 2.3 and 4.2 points, respectively, on the Multi-turn metric. 
Cumulatively, it gets a 6.5 improvement over \textit{SFT} and a 12.5 gain relative to \textit{Base}, highlighting effects in complex, multi-step reasoning tasks.
\begin{figure}[t]
\centering
  \includegraphics[width=0.75\linewidth]{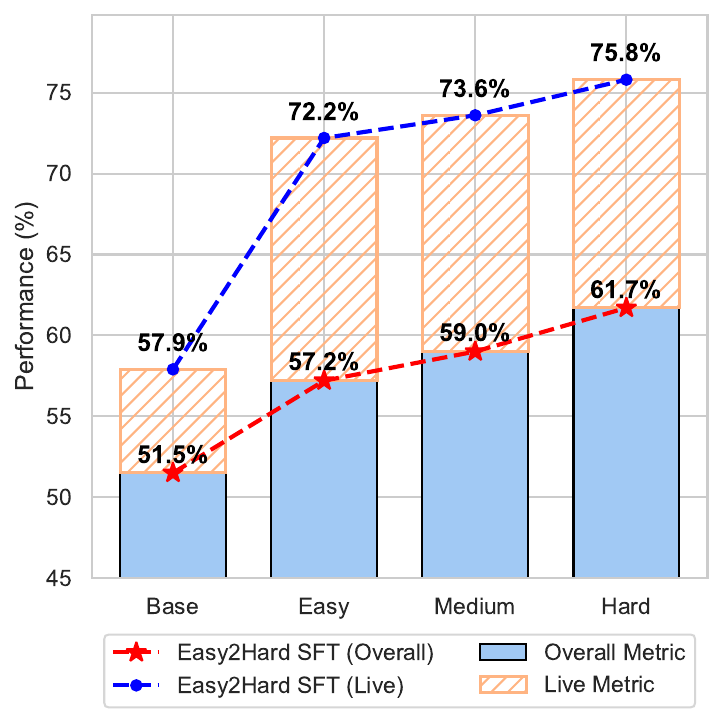}
  \caption {The performance progression of easy to hard warm-up training on \textit{Live} and \textit{Overall} metrics.}
    \label{tab:ablation_sft}
\end{figure}

\begin{figure}[t]
\centering
  \includegraphics[width=0.75\linewidth]{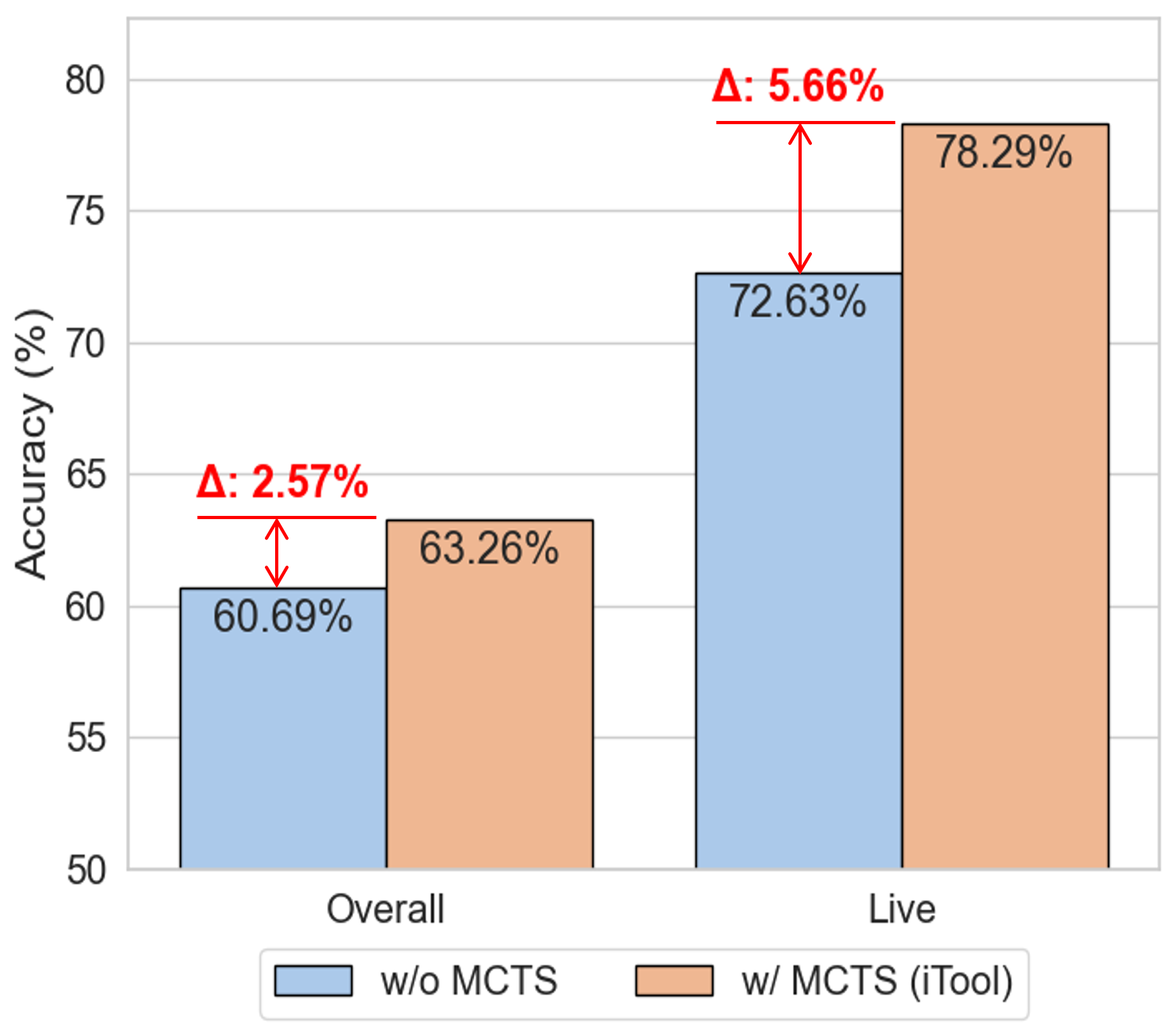}
  \caption {The result of ablation study on MCTS in \textit{iTool} on key metrics.}
    \label{fig:nomcts}
\end{figure}

\subsubsection{Deeper Ablation}
\textbf{(1) In warm-up training, we conducted a study on the easy2hard SFT strategy. }
We present the performance progression from easy to hard and compare it with base model. 
The experimental results are summarized in Figure \ref{tab:ablation_sft}. 
From the results, we observe that our strategy shows a gradual improvement. 
There is a significant leap from \textit{base} to \textit{easy}, and the second largest improvement occurs from the \textit{medium} to \textit{hard}. 
In the synthetic data, the model can quickly learn the task patterns of tool use from the easier stages, which in turn benefits the harder scenario.
This indicates that the model benefits from the curriculum learning process that goes from easy to hard. 

\begin{figure}[th]
\centering
\small
  \includegraphics[width=0.99\linewidth]{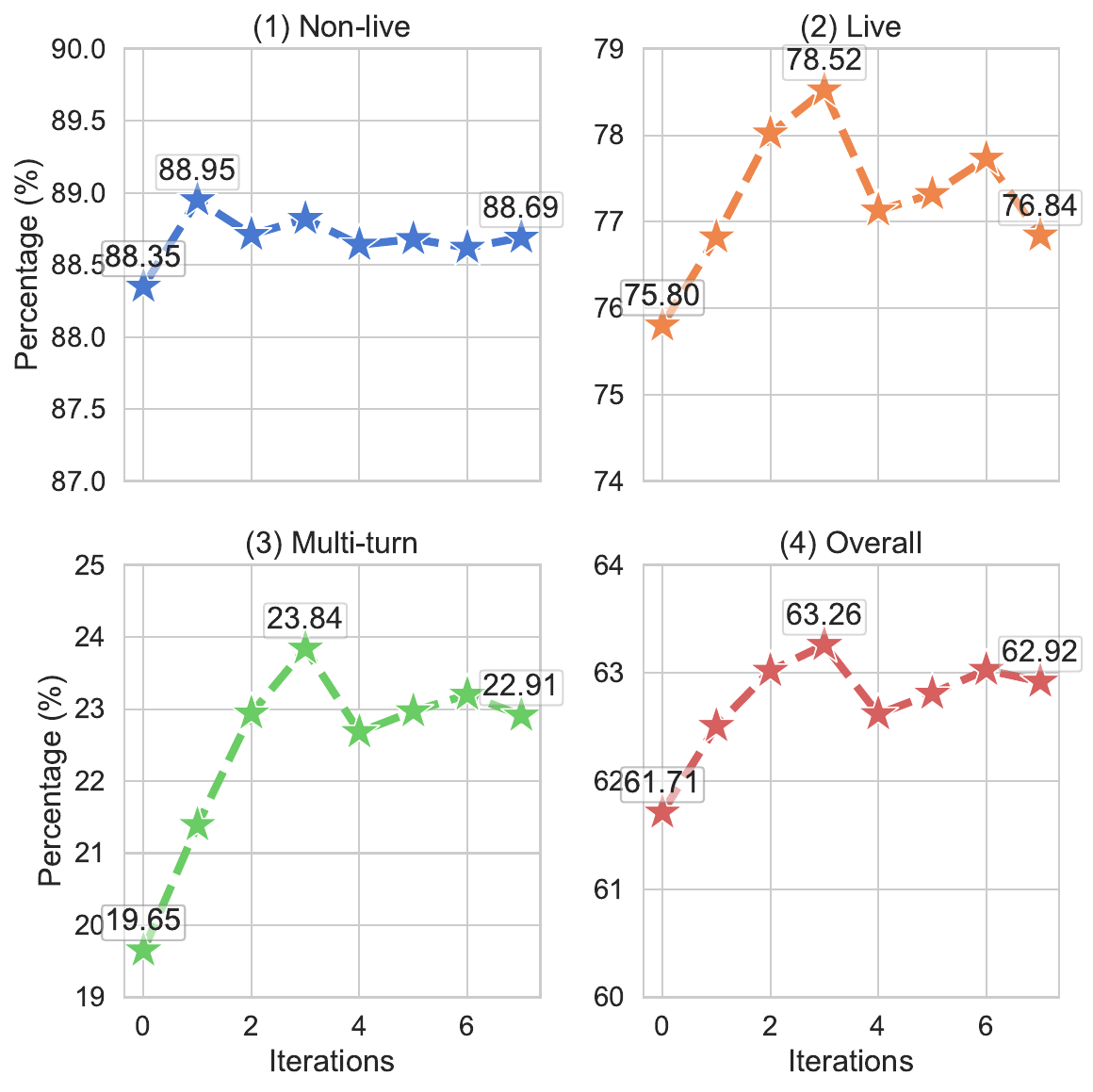}
  \caption {The performance variation of our model with the increase of iterations.}
    \label{fig:ablation_iterations}
\end{figure}

\noindent
\textbf{(2) In iterative reinforcement learning, we conducted a study on MCTS and iteration counts}.
The results are illustrated in Figure \ref{fig:nomcts} and \ref{fig:ablation_iterations} respectively. 
To replace MCTS, we sample four responses from the policy model and select the responses with the highest and lowest probabilities as preference pairs. 
These pairs are then used for subsequent preference optimization (w/o MCTS).
From Figure \ref{fig:nomcts}, we observe that the model's performance deteriorates when MCTS is replaced.
From Figure \ref{fig:ablation_iterations}, we observe that as iterations increase, our method initially shows an upward trend before declining. 
The model performs best around 3 iterations, especially in the Multi-turn and Live scenarios. 
This indicates that MCTS can effectively mitigate the issue of insufficient data diversity with a small number of iterations. 
However, excessive iterations can lead to overfitting, resulting in a decrease in data diversity.

\begin{table*}[th]
    \centering
    \begin{tabular}{lc|cccc|c}
        \toprule
        Base Model & Method & Overall & Non-live & Live & Multi-turn & Rel / Irrel \\
        \hline  \midrule
        \multirow{3}{*}{Llama-3.1-8B-Instruct} & Vanilla & 50.15 &  81.15 &	57.93 &	11.38 &	78.05 /	41.62 \\
                     & Baseline & 60.44 & 88.94 & 74.99 & 17.38 & 80.49 / 85.71 \\
                     & Our & 63.26 & 88.82 & 78.29 & 23.84 & 84.90 / 80.72 \\
        \midrule
        \multirow{3}{*}{Llama-3.2-3B-Instruct}  & Vanilla	& 42.98 & 11.11 &	50.91 &	4.00 &	63.41 /	68.81 \\ 
            & Baseline	& 58.22 & 89.27 & 73.90 & 11.50 &	84.37 / 78.20 \\ 
            & Our	& 62.93 & 90.59 & 76.43 & 15.82 &	84.27 / 87.82 \\ 
        \midrule
        \multirow{3}{*}{Qwen2.5-7B-Instruct}  & Vanilla & 54.27 & 85.58 & 65.97 & 11.25 & 92.68 / 64.95 \\
            & Baseline & 60.69 & 90.02 & 76.23 & 15.92 & 73.47 / 86.98 \\
            & Our & 63.93 & 91.29 & 82.28 & 22.38 & 80.28 / 85.12 \\
        \bottomrule
    \end{tabular}
    \caption{The accuracy performance comparison of base models with different methods on BFCL benchmark. 
     \textit{Vanilla} denotes source base model, \textit{Baseline} denotes supervised fine-tuned base model, \textit{Our} denotes \textit{iTool}.
    }
    \label{tab:base_models}
\end{table*} 

\subsubsection{Base Model Analysis.}
To further validate the effectiveness of base models, we applied our method to other base models.
Due to computational resource constraints, we compared the following base models (\( < 10B\)):
(1) Llama-3.2-3B-Instruct,
(2) Qwen2.5-7B-Instruct \citep{qwen2.5}.
From Table \ref{tab:base_models}, our method exhibits remarkably stable performance across different base models. 
This highlights the robustness of our method in various base models.
On Llama-3.2-3B, our method improved performance by 18\% over the base model. 
On Qwen2.5-7B, it achieved the best performance at 63.22\%.

\subsection{Training Gains Analysis}
\begin{figure}[tb]
\centering
  \includegraphics[width=0.99\linewidth]{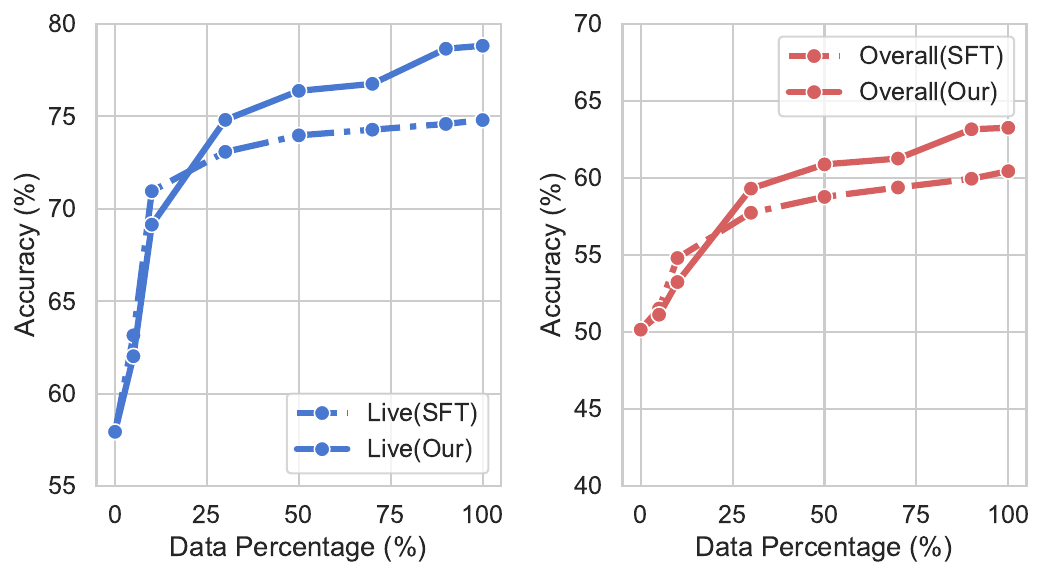}
  \caption {The change curve of training gains as the data scale increases on key metrics.}
    \label{fig:train_gains}
\end{figure}

To analyze the training gains of our method, as detailed in Section \ref{sec:pre_study}, we test the training gains of our method.
From Figure \ref{fig:train_gains}, our method shows greater training gains as the data scale increases in \textit{Live} and \textit{Overall}.
Unlike SFT, whose training benefit curve flattens beyond 30\%, our model exhibits a steeper curve in the Live metric. 
This suggests that our model can alleviate the internal decay of training gains by enhancing its advanced capabilities in complex scenarios.
A additional training cost analysis is conducted in Appendix ~\ref{sec:cost}.


\subsection{Generalization Evaluation of Synthetic Data}
We evaluated the generalization capability of our method across diverse datasets type and model architectures. Experiments included synthetic datasets (Toolace, xLAM\citep{zhang2024xlam}) and a non-synthetic dataset (BFCL-half, using 50\% of BFCL-Live data for training and the remainder for testing). 
Performance was assessed on Llama3.1-8B-Instruct and Llama3.2-3B-Instruct, with results averaged across \textit{Live} and \textit{Multi-turn} metrics.

\begin{table*}[th]
    \centering
    \begin{tabular}{@{}lcccccc@{}}
        \toprule
        \multirow{2}{*}{Dataset (Type)} & \multicolumn{3}{c}{Llama3.1-8B-Instruct} & \multicolumn{3}{c}{Llama3.2-3B-Instruct} \\
        \cmidrule(lr){2-4} \cmidrule(lr){5-7}
        & Baseline (SFT) & iTool & $\Delta$  & Baseline (SFT) & iTool & $\Delta$  \\
        \midrule
        Toolace$\dagger$  & 46.18 & 51.06 & +4.88 & 40.36 & 46.85 & +6.49 \\
        xLAM$\dagger$  & 42.74 & 48.47 & +5.73 & 37.72 & 42.14 & +4.42 \\
        BFCL-half$\ddagger$  & 41.32 & 44.97 & +3.65 & 34.65 & 36.82 & +2.17 \\
        \bottomrule
    \end{tabular}
    \caption{Performance across datasets and models. $\dagger$ denotes synthetic data, and $\ddagger$ denotes non-synthetic data. }
    \label{tab:generalization}
\end{table*}

Our method consistently improved performance across all datasets. 
The largest gains were observed on synthetic datasets (+4.42 to +6.49), with more modest improvements on non-synthetic data (+2.17 to +3.65), demonstrating effective generalization with strongest performance on synthetic benchmarks.
A additional training gain dynamics generalize across model sizes is conducted in Appendix ~\ref{sec:model_scale}.

\section{Related Work}
\subsection{Tool use of LLMs}
Pioneering works like Toolformer \citep{schick2023toolformer} and ToolAlpaca \citep{tang2023toolalpaca} have explored the potential of LLMs in tool use. 
Previously, several tuning-free methods were proposed, which involves manipulating prompts (e.g., \citep{xu2023tool,shi2024learning,qiao2024autoact}) or enhancing execution frameworks (e.g., ReAct \citep{yaoreact}, RestGPT \citep{song2023restgpt}) to unlock inherent capabilities. 

Due to the limitation of user-defined tools in prompts of the above methods, tuning-based methods with synthetic data have been focused.
ToolLlama \citep{toolllm} notably expanded the toolset and investigated the impact of data scaling on performance.
More efficient data synthesis techniques have been proposed for tool use (e.g., ToolACE \citep{liu2024toolace}, BUTTON \citep{chen2024facilitating}, and xLAM \citep{zhang2024xlam}). 

\subsection{Reinforcement Learning}
Learning from human feedback is crucial in aligning LLMs with human intentions \citep{leike2018scalable}, which is known as reinforcement learning.
ReFT enhances this process by combining reinforcement learning with SFT to optimize model performance using reward signals.
Online reinforcement learning algorithms \citep{schulman2017proximal, zheng2023secrets} are complex and difficult to optimize.
Recently, Direct Preference Optimization (DPO) \citep{rafailov2024direct}, a simpler offline algorithm, reparameterizes the reward function to learn a policy model from preference data directly, enhancing simplicity and training stability.
Besides, a variety of preference optimization objectives have been proposed, e.g., SimPo \citep{meng2024simpo}, IPO \citep{azar2024general}, ORPO \citep{hong2024orpo} and KTO \citep{ethayarajh2024kto}.

Further studies have extended this approach to an iterative training setup, by continuously updating the reference model with the most recent policy model or generating new preference pairs at each iteration \citep{dong2024rlhf, yuanself, kim2024sdpo, xiong2024iterRL}

\section{Conclusion}
Equipping LLMs with external tools is becoming a viable method to enhance their capabilities.
In this paper, we study enhancing the advanced tool-use capabilities in a complex scenario from synthetic data.
We find that there are training decay issues when training with synthetic tool-use data.
To alleviate it, we propose an iterative reinforced fine-tuning strategy.
It can continually pinpoint the model's wrong fragments in its responses and address these deficiencies by preference optimization.
The experimental results demonstrate the effectiveness of the proposed method.

\section{Limitation}
While our study has achieved notable advancements, it is important to acknowledge several limitations that could be addressed in future work.
First, the iterative reinforcement learning process (particularly the Monte Carlo Tree Search) requires substantial computational resources to generate fine-grained preference data.
Although it is difficult to solve, we have effectively implemented parameter constraints to manage computational costs efficiently (e.g., 7 hours on 8 V100 GPUs per iteration), achieving a balance between computational feasibility and model performance.
Additionally, due to limited computing resources, we are not able to validate our method on larger 30B or 70B base models. 
Finally, when analyzing the synthetic tool-use data, only a single dataset was tested. 
Testing more publicly available datasets would strengthen the validity and persuasiveness of the conclusions.
We will address these limitations in our future work.

\section*{Acknowledgements}
The research in this article is supported by the New Generation Artificial Intelligence of China (2024YFE0203700), National Natural Science Foundation of China under Grants U22B2059 and 62176079.

\bibliography{custom}

\appendix

\section{Details in Preliminary Study}
\label{sec:error_types}

\subsection{Descriptions of error types}
Here is the descriptions of all error types.
\begin{itemize}
    \item \textbf{Parameter Value}. The value or type of the parameter does not match the ground truth.
    \item \textbf{Parameter Name}. Unable to identify the parameter value from the user query.
    \item \textbf{Parameter Count}. Incorrect number of parameters; required parameters are missing.
    \item \textbf{Tools Count}. The wrong number of tools was called.
    \item \textbf{Tool Name}. There was an error when calling the tool name, such as calling a non-existent tool name or a tool name that does not match the ground truth.
    \item \textbf{Code Syntax}. The tool call does not comply with the syntax of Python, Java, or JavaScript.
    \item \textbf{Other}. Errors other than those mentioned above.
\end{itemize}


\section{Complementary Experiments}
\label{sec:appendix}
\subsection{More Implementation Details}
The experiments were conducted using the publicly available training repository, LLaMA-Factory \citep{zheng2024llamafactory}.
The training of our model can be done within 28 hours with 8 NVIDIA Tesla V100-SXM2-32GB GPUs.
For the training model, we take the best performance checkpoint on the valid dataset.

\noindent
\textbf{The Implementation Settings}. 
Due to resource constraints, we employ a parameter-efficient training strategy using LoRA (with \texttt{rank}=16 and \texttt{alpha}=32) during the SFT warm-up phase, and QLoRA (a quantization method from the bitsandbytes \footnote{https://github.com/TimDettmers/bitsandbytes} library with 4 bits) during the reinforcement learning (RL) phase.
We utilize a cosine learning rate scheduler with a warm-up ratio of 0.1.
More detailed training settings are shown in Table \ref{tab:train_set}.
\begin{table}[h]
    \centering
        \begin{tabular}{l|clc}
        \toprule
        {Stage} & {epoch} & {lr} & {batch size} \\ \hline
        \multirow{3}{*}{SFT} & \multirow{3}{*}{3} & easy: 5e-5 & \multirow{3}{*}{64} \\ 
         &  &  medium: 2e-5 &   \\ 
         &  &  hard: 1e-5 & \\ \midrule
        {RL} & 2 & 1e-6 & 64  \\ 
        \bottomrule
        \end{tabular}
    \caption{ The detailed training settings in our method. 
    \texttt{lr} denotes learning rate. \texttt{batch size} denotes the total batch size, equals 1 (per device) times 8 (accumulation steps) times 8 (devices). }
    \label{tab:train_set}
\end{table}

\textbf{Implementation Settings in MCTS-base RL}. 
In \textit{Expand} phase of MCTS, the prompt for self-evaluation is shown in Table \ref{eval_prompt}.
When calculating the confidence score for correctness, we evaluate the token-level probabilities of a policy model across four options (\texttt{A, B, C, D}) with respective weights of 1.0, 0.1, -1.0, and -2.0. 
We sample the model's responses four times and use the weighted average of these samples as the final confidence score.

To ensure the quality of the sampled preference data, we exclude the following data: (1) pairs with candidate step similarity above 95\%, (2) pairs with a \( \mathcal{Q} \)-value difference less than 0.1, and (3) accepted samples with a \( \mathcal{Q} \)-value below 0.3.
In MCTS, to control algorithm overhead, we limit the following parameters: (1) \texttt{depth}, the maximum depth of the search tree, (2) \texttt{width}, the maximum number of child nodes per node, (3) \texttt{simulation}, the maximum number of simulation steps in \textit{Expand} phase, and (4) \texttt{iterations}, the maximum number of iterations to construct the MCTS search tree.
We summarize these parameters in Table \ref{tab:parameters}.

\begin{table}[th]
    \centering
    \begin{tabular}{lc|lc} \\ \toprule
        {Parameters} & {Value} & {Parameters} & {Value} \\ \midrule
        \texttt{depth} & 3 & $c$ & 1.0 \\ 
        \texttt{width} & 3 & \texttt{temperature} & 1.5 \\
        \texttt{simulation} & 2 & \texttt{seed} & 42 \\
        \texttt{iterations} & 5 &  &  \\ 
    \bottomrule
    \end{tabular}
    \caption{The parameters setting in MCTS. $c$ denotes the degree of exploration in the \textit{Select} phase.}
    \label{tab:parameters}
\end{table}

\subsection{Cost Analysis}
\label{sec:cost}
We conducted a cost-benefit analysis to evaluate iTool's performance gains against computational overhead, focusing on MCTS sampling efficiency. 
Experiments compared the base model, SFT baseline, and iTool across accuracy metrics (BFCL-Live and Multi-turn) and time costs, using an 8×32G V100 GPU configuration.

\begin{table}[th]
  \centering
  \small
  \begin{tabular}{@{}lccc@{}}
    \toprule
    Model       & Live  & Multi-turn & Time Cost \\
    \midrule
    Base Model  & 57.93         & 11.38               & 0h        \\
    SFT Baseline& 74.99         & 17.38               & 10h       \\
    iTool       & 78.29 \textcolor{green!40!black}{{↑3.3}} & 23.84 \textcolor{green!40!black}{{↑6.46}}       & 28h ($\times$2.8) \\
    \bottomrule
  \end{tabular}
  \caption{Cost-benefit analysis of different models}
  \label{tab:cost_analysis}
\end{table}

Results in Figure~\ref{tab:cost_analysis} show iTool outperforms the SFT baseline by 3.30\% in BFCL-Live accuracy and 6.46\% in Multi-turn accuracy, with a 2.8× increase in time cost. 
The significant gains in complex Multi-turn scenarios, where complexity is highest, demonstrate favorable cost-effectiveness for practical deployment.

\subsection{Generalize Across Model Sizes}
\label{sec:model_scale}
\begin{figure}[th]
\centering
  \includegraphics[width=0.8\linewidth]{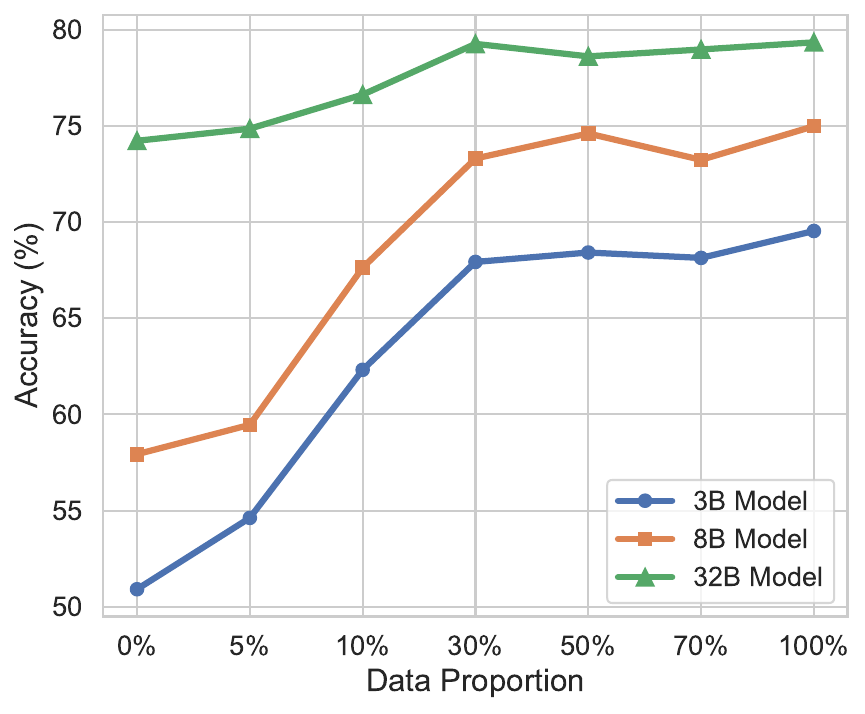}
  \caption {Training gain dynamics generalize across model sizes.}
    \label{fig:model_scale}
\end{figure}

To investigate the efficacy of SFT at scale and examine whether training gain dynamics generalize across model sizes, we conducted a controlled SFT study using three open-source instruction-tuned models of increasing capacity: Llama3.2-3B-Instruct, Llama3.1-8B-Instruct, and Qwen2.5-32B-Instruct. 
Each model was fine-tuned on incrementally scaled subsets of training data, ranging from minimal to full data regimes. 
Performance was evaluated on the \textit{BFCL-Live} benchmark to track accuracy progression as a function of data volume, as shown in Figure ~\ref{fig:model_scale}. 
The results demonstrate that, across all three model scales, the marginal gains from additional training data follow a decaying trend, that is, performance improvements diminish as data scale increases, indicating consistent saturation behavior regardless of model size. 
This suggests that while larger models achieve higher absolute performance, their relative gains from scaling data during SFT exhibit predictable attenuation, reinforcing the importance of data efficiency strategies even at large scales.

\subsection{Preference Algorithm Analysis}
{ In iterative reinforcement learning, we also explore different preference optimization algorithms.}
Besides the widely used DPO \citep{rafailov2024direct}, we also explored SimPO \citep{meng2024simpo}, IPO \citep{azar2024general}, and ORPO \citep{hong2024orpo}. 
DPO reparameterizes the reward function to learn a policy model from preference data directly.
IPO is a theoretically grounded approach method that avoids DPO’s assumption that pairwise
preferences can be replaced with pointwise rewards.
ORPO introduces a reference-model-free odd ratio term to directly contrast winning and losing
responses with the policy model and jointly trains with the SFT objective.
SimPO aligns the reference-free reward function in the preference optimization objective with the generation metric.
For fair comparisons, we start these algorithms from the same SFT checkpoints, the reference model is initialized as the policy model.

\begin{figure}[th]
\centering
  \includegraphics[width=0.85\linewidth]{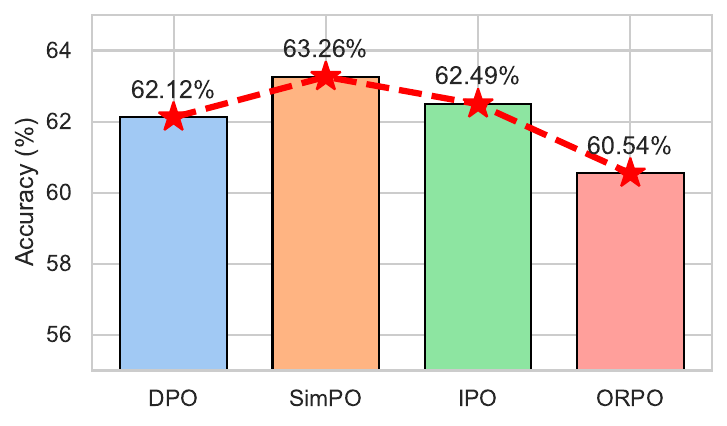}
  \caption {The performance \textit{iTool} using different preference optimization algorithms on BFCL.}
    \label{fig:dpo}
\end{figure}

For these algorithms, we conducted a thorough search for the optimal hyperparameter settings to ensure a fair comparison. 
The results of hyperparameter settings are shown in Table \ref{tab:po}.
The results of different preference optimization algorithm with optimal hyperparameter settings are shown in Figure \ref {fig:dpo}.
From the result, we find \textit{iTool} with SimDPO achieved the best performance.
Different preference algorithms do not create significant performance gaps except for ORPO.

\begin{table*}[h]
    \small
    \centering
        \begin{tabular}{lll|l}
        \hline
        \textbf{Method} & \textbf{Objective} & \textbf{Hyperparameters} & \textbf{Best Setting} \\ \hline
        
        \multirow{2}{*}{DPO}  &\multirow{2}{*}{ $-\log \sigma \left(\beta \log \frac{\pi_{\theta}(y_w|x)}{\pi_{\text{ref}}(y_w|x)} - \beta \log \frac{\pi_{\theta}(y_l|x)}{\pi_{\text{ref}}(y_l|x)}\right)$ } & $\beta \in [0.01, 0.05, 0.1]$ & $\beta = 0.1$\\
         & & $lr \in [1e-6, 5e-7, 3e-7]$ & $lr = 3e-7$ \\ \hline
        
        \multirow{2}{*}{IPO}   & \multirow{2}{*}{$\left(\log \frac{\pi_{\theta}(y_w|x)}{\pi_{\text{ref}}(y_w|x)} - \log \frac{\pi_{\theta}(y_l|x)}{\pi_{\text{ref}}(y_l|x)} - \frac{1}{2\tau}\right)^2$ } & $\tau \in [0.01, 0.05, 0.1]$ & $\tau = 0.1$ \\
         & & $lr \in [1e-6, 5e-7, 3e-7]$ & $ lr = 1e-6$ \\ \hline
        
        \multirow{2}{*}{ORPO}  & $-\log p_{\theta}(y_w|x) - \lambda \log \sigma \left(\log \frac{p_{\theta}(y_w|x)}{1 - p_{\theta}(y_w|x)} - \log \frac{p_{\theta}(y_l|x)}{1 - p_{\theta}(y_l|x)}\right)$, & $\lambda \in [0.01, 0.05, 0.1]$ & $\lambda = 0.1$ \\
         & where $p_{\theta}(y|x) = \exp\left(\frac{1}{|y|} \log \pi_{\theta}(y|x)\right)$ &  $lr \in [1e-6, 5e-7, 3e-7]$ & $lr = 3e-7$ \\ \hline
        
        \multirow{3}{*}{SimPO} & \multirow{3}{*}{$-\log \sigma \left(\frac{\beta}{|y_w|} \log \pi_{\theta}(y_w|x) - \frac{\beta}{|y_l|} \log \pi_{\theta}(y_l|x) - \gamma\right)$ }& $\beta \in [2.0, 2.5]$ & $\beta = 2.5$\\
         & & $\gamma \in [0.5, 1.0, 1.4]$ & $\gamma = 0.5$ \\
         & & $lr \in [1e-6, 5e-7, 3e-7]$ & $lr = 1e-6$ \\ \hline
        \end{tabular}
    \caption{The search for optimal hyperparameter settings of different preference optimization algorithms.}
    \label{tab:po}
\end{table*}

\begin{table*}[ht]
    \centering
    \begin{tcolorbox}[colback=gray!10!white, colframe=black, title=Prompt 1: Eval Prompt, fonttitle=\bfseries, label=prompt2, width=0.9\textwidth, left=1mm, right=1mm, center title]
    \small
    Ground Truth Response: \texttt{\{gt\_ans\}} \\
    Generated Response by Model: \texttt{\{response\}} \\
    
    \vspace{2mm}
    User Instruction: \\
    Please assess the quality of the generated response relative to the ground truth response. \\
    Note: A generated response that is a fragment of the ground truth response is also excellent. \\
    
    \vspace{2mm}
    Evaluation Criteria: \\
    1. Function Name: Is the name of all the function called correct? \\
    2. Parameter Count: Is the number of parameters for all the function correct? \\
    3. Parameter Names: Are the names of all the parameters for the function correct? \\
    4. Parameter Value/Types: Are the value/types of all the parameters for the function correct? \\
    5. Semantic Similarity: Is the generated response semantically close to the ground truth response? \\
    
    \vspace{2mm}
    Please directly choose from the following options to judge the overall quality: \\
    (A) Excellent: The generated response meets all criteria and is almost identical to the ground truth response. \\
    (B) Acceptable: The generated response meets most criteria but has minor discrepancies. \\
    (C) Fair: The generated response meets some criteria but has significant issues. \\
    (D) Poor: The generated response fails to meet most or all criteria. \\
    
    \vspace{2mm}
    ASSISTANT: The option of overall quality is \\
    You are an AI specialized in tool use. \\
    Your task is to assess the potential veracity of \texttt{\{placeholder\}}.
    \end{tcolorbox}
    \caption{The Eval Prompt for self-evaluation in Eq. \ref{eval} of Section \ref{sec:method}.}
    \label{eval_prompt}
\end{table*}

\section{Case Analysis}

\subsection{An Example of Preference Pair}
\label{example_pair}
Table \ref{tab:case_pair1} illustrates a preference pair example. 
The \texttt{chosen} response correctly employs the "Get Trending Result" tool with suitable parameters for the user's request. 
Conversely, the \texttt{rejected} response is improperly formatted, omits necessary parentheses, and incorrectly assigns the value \texttt{1} to the \texttt{timeframe} parameter, showcasing an erroneous application of the tool.

\begin{figure*}[th]
\centering
  \includegraphics[width=0.9\linewidth]{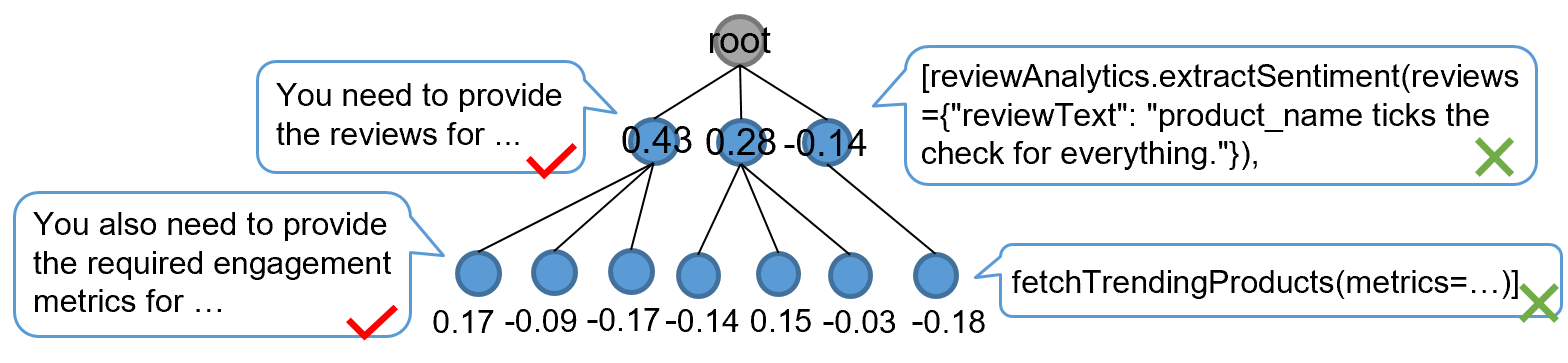}
  \caption {The illustration of example 2 in Table \ref{tab:case_pair2} for preference pair derived from MCTS. 
 The floating-point values of nodes denote the \( \mathcal{Q} \)-value in MCTS.}
    \label{fig:case_pair2}
\end{figure*}

Table \ref{tab:case_pair2} presents another case of preference pair, sampled during the MCTS research tree as depicted in Figure \ref{fig:case_pair2}. 
In this scenario, the user's query lacks the specific details necessary for the functions mentioned (i.e., \texttt{reviews} for '{reviewAnalytics.extractSentiment}' and \texttt{metrics} for '{socialTrends.fetchTrendingProducts}').
The assistant's chosen response correctly identifies the need for these parameter values, whereas the rejected response incorrectly hallucinates when recognizing these parameters.

\begin{table*}[th]
    \small
    \centering
    \begin{tabular}{>{\ttfamily}lp{0.85\linewidth}}
        \toprule
        \multicolumn{2}{c}{Example 1} \\ \midrule
        Tool list &  \begin{minipage}[t]{\linewidth}
\begin{verbatim}
[{
    "name": "Get Trending Result",
    "description": "Retrieves trending search results from Google Trends for 
        the United States.",
    "parameters": {
      "type": "dict",
      "properties": {
        "category": {
          "description": "The category of trending results (e.g., news, entertainment, etc.)",
          "type": "string"
        },
        "timeframe": {
          "description": "The timeframe for which to retrieve trending results (e.g., today, 
            this week, etc.)",
          "type": "string"
        }
      },
      "required": ["category"]
    },
    "required": null
},{
    "name": "Get Server Time",
    "description": "Returns the current server time in a standardized format",
    "parameters": {
      "type": "dict",
      "properties": {},
      "required": []
    },
    "required": null
}] \end{verbatim} 
        \end{minipage}  \\ \midrule
        User & Can you tell me the latest trending topics in entertainment today? \\ \midrule
        \multicolumn{2}{l}{Assistant} \\ \midrule
        chosen & [Get Trending Result(category="entertainment", timeframe="today")] \\ 
        rejected & [Get Trending Result category="entertainment", timeframe="1"] \\  \bottomrule
    \end{tabular}
    \caption{ The example 1 of preference pair derived from MCTS.}
    \label{tab:case_pair1}
\end{table*}

\begin{table*}[h]
    \small
    \centering
    \begin{tabular}{>{\ttfamily}lp{0.85\linewidth}}
        \toprule
        \multicolumn{2}{c}{Example 2} \\ \midrule
        Tool list &  \begin{minipage}[t]{\linewidth}
\begin{verbatim}
[{
    "name": "fetchTrendingProducts",
    "description": "Fetches trending products based on reviews and engagement metrics.",
    "parameters": {
      "metrics": {
        "type": "array",
        "items": {
          "productID": "string",
          "engagementScore": {
            "likes": "int",
            "shares": "int"
    } } } }
},{
    "name": "extractSentiment",
    "description": "Extracts sentiment scores from reviews.",
    "parameters": {
      "reviews": {
        "type": "array",
        "items": {
          "reviewText": "string",
          "reviewDate": {
            "type": "string",
            "pattern": "^\\w+ \\d{1,2}, \\d{4}$"
    } } } }
}] \end{verbatim} 
    \end{minipage}  \\ \midrule
        User & I'm running a marketing campaign and need sentiment analysis from this month's reviews of our main product. 
        We had 300 likes and 50 shares yesterday. 
        Please analyze trends and sentiment for our competitor's product too; they had 450 likes and 75 shares. \\ \hline
        \multicolumn{2}{l}{Assistant} \\ \hline
        chosen & \texttt{You need to provide the reviews for the sentiment analysis of the flagship product and the competitor's product.} \\ 
        rejected & \texttt{[reviewAnalytics.extractSentiment(reviews=\{"reviewText": "product\_name ticks the check for everything."\}),} \\ \midrule
    \end{tabular}
    \caption{ The example 2 of preference pair derived from MCTS.}
    \label{tab:case_pair2}
\end{table*}

\end{document}